%% file: main.tex
\pdfoutput=1
\documentclass[a4paper,twoside]{article}

\usepackage{epsfig}
\usepackage[utf8]{inputenc} %
\usepackage[T1]{fontenc}    %
\usepackage{hyperref}       %
\usepackage{url}            %
\usepackage{booktabs}       %
\usepackage{amsfonts}       %
\usepackage{nicefrac}       %
\usepackage{microtype}      %
\usepackage{xcolor}         %

\usepackage{graphicx} %
\usepackage{amsmath}
\usepackage{xcolor}
\usepackage{comment}
\usepackage{multirow}
\usepackage{subcaption}
\usepackage{calc}
\usepackage[ruled,vlined,linesnumbered]{algorithm2e}
\usepackage{cases}
\usepackage{multicol}
\usepackage{tabularx}
\usepackage{dsfont}
\usepackage{nicefrac}       %
\usepackage{amsthm}
\usepackage{amstext}
\usepackage{pslatex}
\usepackage{microtype}      %
\usepackage{xcolor}         %
\usepackage{amssymb}
\usepackage{apalike}
\usepackage{tikz}
\usepackage{pgfplots}
\usepackage[bottom]{footmisc}
\usepackage{bm}

\newcolumntype{Y}{>{\centering\arraybackslash}X}

\SetKwInput{KwInput}{Input}
\SetKwInput{KwOutput}{Output}

\usepackage{thmtools}

\theoremstyle{definition}
\newtheorem{definition}{Definition}

\theoremstyle{remark}
\newtheorem*{remark}{Remark}

\definecolor{britishracinggreen}{rgb}{0.0, 0.26, 0.15}

\usepackage{soul}

\usepackage{SCITEPRESS}
\begin{document}

\title{Data Augmentation through Expert-guided Symmetry Detection to Improve Performance in Offline Reinforcement Learning}

\author{\authorname{Giorgio Angelotti\sup{1,2}\orcidAuthor{0000-0002-1878-5833}, Nicolas Drougard\sup{1,2}\orcidAuthor{0000-0003-0002-9973}
and Caroline P. C. Chanel\sup{1,2}\orcidAuthor{0000-0003-3578-4186}
}
\affiliation{\sup{1}ISAE-SUPAERO, University of Toulouse, France}
\affiliation{\sup{2}ANITI, University of Toulouse, France}
\email{\{name.surname\}@isae-supaero.fr}
}


\abstract{Offline estimation of the dynamical model of a Markov Decision Process (MDP) is a non-trivial task that greatly depends on the data available in the learning phase. Sometimes the dynamics of the model is invariant with respect to some transformations of the current state and action. Recent works showed that an expert-guided pipeline relying on Density Estimation methods as Deep Neural Network based Normalizing Flows effectively detects this structure in deterministic environments, both categorical and continuous-valued. The acquired knowledge can be exploited to augment the original data set, leading eventually to a reduction in the distributional shift between the true and the learned model. Such data augmentation technique can be exploited as a preliminary process to be executed before adopting an Offline Reinforcement Learning architecture, increasing its performance. In this work we extend the paradigm to also tackle non-deterministic MDPs, in particular, 1) we propose a detection threshold in categorical environments based on statistical distances, and 2) we show that the former results lead to a performance improvement when solving the learned MDP and then applying the optimized policy in the real environment.}

\onecolumn \maketitle \normalsize \setcounter{footnote}{0} \vfill
Preprint. Accepted at ICAART 2023\footnote{Preprint. Accepted at ICAART 2023}.
\input{sections/0-introduction}
\input{sections/1-background}
\input{sections/2-improvements}
\input{sections/3-experiments}
\input{sections/4-discussion}

\input{sections/5-conclusion}

\section*{Acknowledgements}
This work was funded by the Artificial and Natural Intelligence Toulouse Institute (ANITI) - Institut 3iA (ANR-19-PI3A-0004).

\bibliographystyle{apalike}
{\small
\bibliography{references}}
\end{document}

%% file: sections/0-introduction.tex
\section{Introduction}
In Offline Reinforcement Learning (ORL) and Offline Learning for Planning the environment dynamics and/or value functions are inferred from a batch of already pre-collected experiences. Wrong previsions lead to bad decisions. The distributional shift, defined as the discrepancy between the learnt model and reality, is the main responsible for the performance deficit of the (sub)optimal policy obtained in the offline setting compared to the true optimal policy \cite{Levine2020OfflineRL,angelotti2020offline}.
Is there a way to exploit expert knowledge or intuition about the environment to limit the distributional shift?
Several models benefit from a dynamics that is invariant with respect to some transformations of the system of reference. In physics, such a property of a system is called a symmetry \cite{Gross14256}. In the context of Markov Decision Processes (MDPs) \cite{bellman1966dynamic} a symmetry can be defined as a particular case of an MDP's homomorphism \cite{icart}.
Knowing that a system to be learned is endowed with a symmetry or of a homomorphic structure can lead to more data-efficient solutions of an MDP.

The automatic discovery of homomorphic structures in MDPs has a long story \cite{dean1997model,ravindran2001symmetries,ravindran2004approximate}. In \cite{li2006towards} a theoretical analysis of the possible types of MDPs state abstractions proved which properties of the original MDP would be invariant under the transformation: the optimal value function, the optimal policy, etc. Eventually, the full automatic discovery of a factored MDP representation was proven to be as hard as verifying whether two graphs are isomorphic \cite{narayanamurthy2008hardness}.
In recent years \cite{10.5555/3398761.3398926,NEURIPS2020_2be5f9c2,icart} rekindled the topic. 

In \cite{10.5555/3398761.3398926} 
a contrastive loss function that enforces action equivariance on a to-be-learned representation of an MDP was adopted to learn a structured latent space that was then exploited to increase the data efficiency of a data-driven planner. \cite{NEURIPS2020_2be5f9c2} introduced peculiar classes of Deep Neural Network (DNN) architectures that by construction enforce the invariance of the optimal MDP policy under some set of transformations obtained through other Deep RL paradigms. The latter also provided an increase in data efficiency.
In \cite{icart} an expert-guided detection of alleged symmetries based on Density Estimation statistical techniques in the context of the offline learning of both continuous and categorical environments was proposed in order to eventually augment the starting data set. The authors showed that correctly detecting a symmetry (based on the computation of a symmetry confidence value $\nu_k > \nu$) and data augmenting the starting data set exploiting this information led to a decrease in the distributional shift. Unfortunately, the said work concerned only \textit{deterministic} MDPs and did not include an analysis of the \textit{performance} of the policy obtained in the end. In other fields of Machine Learning data augmentation has been extensively exploited to boost the efficiency of the algorithms in data-limited setups \cite{dataaug,shorten2019survey,park2019specaugment}.

Recently \cite{yarats2022don} showed the importance of large and diverse datasets for ORL by demonstrating empirically that offline learning using a vanilla online RL algorithm over a batch that is diverse enough can lead to performances that are comparable to, or even better than, pure ORL approaches.

In this context, the present work addresses the following research questions:
\textit{Is it possible to develop a method for expert-guided detection of alleged symmetries based on Density Estimation statistical techniques in the context of offline learning that also works for stochastic MDPs?} The main idea is to extend previous works \cite{10.5555/3398761.3398926,NEURIPS2020_2be5f9c2,icart} to deal with stochastic MDPs; and, 
\textit{Is Data Augmentation exploiting a detected symmetry really beneficial to the learning of an MDP policy in the offline context?} We would like to empirically demonstrate (O)RL policy improvement when enriching the batch as proposed by \cite{yarats2022don}.

\paragraph{Contributions}
In this work, we take over and extend the state-of-the-art with the aim of providing an answer to the listed research questions.
More specifically, the contributions of this paper are the followings:
\begin{enumerate}
    \item \textit{Algorithmic contribution:} A refinement of the decision threshold, based on statistical distances, is defined for categorical MDPs. This new decision threshold is valid also in both stochastic and deterministic environments, improving hence over the state-of-the-art that only tackled deterministic scenarios; 
    \item \textit{Experimental contribution:} The improvement of the policy performance obtained by augmenting the data with the symmetric images of the transitions is demonstrated experimentally in an offline learning context. The good quality of the method is clear in the categorical setting while it is fuzzier in the continuous setting since offline methods with Deep Neural Networks are affected by the (non-trivial) choice of the hyperparameters.
\end{enumerate}
It is worth saying that the presented work's aim is not to be a competitor to the ORL algorithms, but a way to augment the batch by validating expert intuition. Once the batch has been augmented one could use any offline RL method.

%% file: sections/1-background.tex
\section{Background}
\begin{definition}[Markov Decision Process]
An MDP \cite{bellman1966dynamic} is a tuple $\mathcal{M} = (S, A, R, T, \gamma)$. $S$ and $A$ are the sets of states and actions, $R : S \times A \rightarrow \mathbb{R}$ is the
reward function, $T : S \times A \rightarrow Dist(S)$ is the transition function, where $Dist(S)$ is the set of probability distributions on $S$, and $\gamma \in [0, 1)$ is the discount factor. Time is discretized and at each step $t \in \mathbb{N}$ the agent observes a system state $s=s_t \in S$, acts with $a=a_t \in A$ drawn from a policy $\pi : S  \rightarrow Dist(A)$, and with probability $T(s, a, s')$ transits to a next state $s'=s_{t+1}$, earning a reward $R(s, a)$. The value function of $\pi$ and $s$ is defined as the expected total discounted reward using $\pi$ and starting with $s$: $V_{\pi}(s) = \mathbb{E}_{\pi}\big[\sum_{t=0}^{\infty}\gamma^{t} R(s_t, a_t)|s_0 = s\big]$. The optimal value function $V^{*}$ is the maximum of the latter over every policy $\pi$.
\end{definition}
\begin{definition}[MDP Symmetry]
Given an MDP $\mathcal{M}$, let $k$ be a surjection 
on $S \times A \times S$ such that $k(s,a,s')=\big(k_{\sigma}(s,a,s'),k_{\alpha}(s,a,s'),k_{\sigma'}(s,a,s')\big) \in S \times A \times S$.
Let $(T\circ k)(s,a,s'\big) = T\left(k(s,a,s')\right)$.
$k$ is a symmetry if $\forall (s,s')\in S^2$, $a \in A$ both $T$ and $R$ are invariant with respect to the image of $k$:
\begin{equation}
\label{eq:dyn}
(T\circ k)(s,a,s'\big) = T(s,a,s'),
\end{equation}
\begin{equation}
R\big(k_{\sigma}(s,a,s'),k_{\alpha}(s,a,s')\big) = R(s,a).
\end{equation}
\end{definition}
As \cite{icart}, in this paper we will focus only on the invariance of $T$, therefore we will only demand for the validity of Equation \ref{eq:dyn}. Problems with a known reward function as well as model-based approaches can thus benefit directly from the method.
\paragraph{Probability Mass Function Estimation for Discrete MDPs.}
Let $\mathcal{D} = \{(s_i, a_i, s_i')\}_{i=1}^n$ be a batch of recorded transitions. Performing mass estimation over $\mathcal{D}$ amounts to compute the probabilities that define the categorical distribution $T$ by estimating the frequencies of transition in $\mathcal{D}$.
In other words: 
\begin{equation}
\label{eq:discrt}
    \hat{T}(s,a,s') = \begin{cases}
    \dfrac{n_{s,a,s'}}{\sum_{s'} n_{s,a,s'}} &\text{if $\sum_{s'} n_{s,a,s'} > 0$,}\\
|S|^{-1} &\text{otherwise.}\end{cases}
\end{equation}
where $n_{s,a,s'}$ is the number of times the transition $(s_t = s, a_t = a, s_{t+1} = s') $ appears in $\mathcal{D}$.

\paragraph{Probability Density Function Estimation for Continuous MDPs.}
Performing density estimation over $\mathcal{D}$ means obtaining an analytical expression for the probability density function (pdf) of transitions $(s, a, s')$ given $\mathcal{D}$: $\mathcal{L}(s, a, s' | \mathcal{D})$. Normalizing flows \cite{DBLP:journals/corr/DinhKB14,kobyzev2020normalizing} allow defining a parametric flow of continuous transformations that reshapes a known initial pdf to one that best fits the data.
\paragraph{Expert-guided detection of symmetries}
The paradigm described in \cite{icart} can be resumed as follows:
\begin{enumerate}
    \item An expert presumes that a to be learned model is endowed with the invariance of $T$ with respect to a transformation $k$;
    \item She/He computes the probability function estimation based on the batch $\mathcal{D}$:
    \begin{enumerate}
        \item (categorical case) She/He computes $\hat{T}$, an estimate of $T$, using the transitions in a batch $\mathcal{D}$ by applying Equation \ref{eq:discrt};
        \item (continuous case) She/He performs Density Estimation over $\mathcal{D}$ using Normalizing Flows;
    \end{enumerate}
    \item She/He applies $k$ to all transitions $(s,a,s') \in \mathcal{D}$ and then checks whether the symmetry confidence value $\nu_k$; %
    \begin{enumerate}
        \item (categorical case) of samples $k(s,a,s') = \big(k_{\sigma}(s,a,s'), k_{\alpha}(s,a,s'), k_{\sigma'}(s,a,s')\big) \in k(\mathcal{D})$ s.t. $T(s,a,s') = (T\circ k) (s,a,s')$ exceeds an expert given threshold $\nu$;
        \item (continuous case) of probability values $\mathcal{L}$ evaluated on $k(\mathcal{D})$ exceeds a threshold $\theta$ that corresponds to the $q-$order quantile of the distribution of probability values evaluated on the original batch. The quantile order $q$ is given as an input to the procedure by an expert (see Algorithm \ref{algo:cont});
    \end{enumerate}
    \item If the last condition is fulfilled then $\mathcal{D}$ is augmented with $k(\mathcal{D})$.
\end{enumerate}
Note that once a transformation $k$ is detected as a symmetry the dataset is potentially augmented with transitions that are not present in the original batch, injecting hence unseen and totally novel information into the dataset.

%% file: sections/2-improvements.tex
\section{Algorithmic Contribution}
Our algorithmic contribution consists in the improvement of the calculation of $\nu_k$ in part (3.a) of the previous list \cite{icart}. Indeed, that approach does not yield valid results when applied to stochastic environments. In order for the method to work in stochastic environments we need to measure a distance in distribution. The latter somehow was considered in the version of the approach that took care of continuous deterministic environments since learning a distribution over transitions represented by their features is independent of the nature of the dynamics. However, when dealing with categorical states the notion of distance between features can't be exploited.

We propose to compute the percentage $\nu_k$ relying on a distance between categorical distributions. Since the transformation $k$ is a surjection on transition tuples, we do not know a-priori which will be the correct mapping $k_{\sigma'}(s,a,s')$ $\forall s' \in S$. In other words, we can compute $k_{\sigma'}$, the symmetric image of $s'$, only when we receive as an input the whole tuple $(s,a,s')$ since an inverse mapping might not exist.

Therefore we will resort to computing a \emph{pessimistic} approximation of the Total Variational Distance (proportional to the $L^1$-norm). In particular, given $(s,a,s')$, we aim to calculate the Chebyshev distance (the $L^{\infty}$-norm) between $T(s,a,\cdot)$ and $T\big(k_{\sigma}(s,a,s'),k_{\alpha}(s,a,s'),\cdot\big)$. Recall that given two vectors of dimension $d$, $x$ and $y$ both $\in \mathbb{R}^d$, $\lvert \lvert x-y \rvert \rvert_{\infty}\leq \lvert \lvert x-y \rvert \rvert_{1}$.

Let us then define the following four functions:
\begin{equation}
\label{eq:lilm}
m(s,a,s') = \min_{\overline{s} \in S\setminus\{s'\}:\hat{T}\neq 0}{\hat{T}(s,a,\overline{s})}
\end{equation}
\begin{equation}
\label{eq:capm}
M(s,a,s') = \max_{\overline{s} \in S\setminus\{s'\}}{\hat{T}(s,a,\overline{s})},
\end{equation}
\begin{equation}
\label{eq:lilmk}
\resizebox{\columnwidth}{!}{%
$m_k(s,a,s') = \min_{ \substack{\overline{s} \in S \mbox{ \tiny s.t.} \\ 
    \overline{s} \neq k_{\sigma'}(s,a,s') \\ \mbox{\tiny and } \hat{T} \circ k \neq 0}}  \hat{T}\big(k_{\sigma}(s,a,s'),k_{\alpha}(s,a,s'),\overline{s}\big)$},
\end{equation}
\begin{equation}
\label{eq:capmk}
\resizebox{\columnwidth}{!}{%
$M_k(s,a,s') = \max_{ \substack{\overline{s} \in S \mbox{ \tiny s.t.} \\ \overline{s} \neq k_{\sigma'}(s,a,s')} }{\hat{T}\big(k_{\sigma}(s,a,s'),k_{\alpha}(s,a,s'),\overline{s}\big)}$}
\end{equation}
where $m$ ($M$) and $m_k$ ($M_k$) are the minimum (maximum) of the probability mass function (pmf) %
$\hat{T}$ when evaluated respectively on an initial state and action $(s,a)$ and $\big(k_{\sigma}(s,a,s'),k_{\alpha}(s,a,s')\big)$ for which $\hat{T}\neq 0$. Those zero values are excluded because, in the context of a small dataset, many transitions are unexplored, and including values $= 0$ would often lead to over-pessimistic estimates.

In order to approximate the Chebyshev distance between $\hat{T}\left(s,a,\cdot\right)$ and $\hat{T}\left(k_{\sigma}(s,a,s'), k_{\alpha}(s,a,s'),\cdot\right)$ we define a pessimistic approximation $d_k$ as follows:
\begin{align}
\nonumber d_k(s,a,s') = \max & \Big\{ \underbrace{\big|M(s,a,s')-m_k(s,a,s')\big|}_{(I)}, \\ & \label{eq:distance}\underbrace{\big|M_k(s,a,s') - m(s,a,s')\big|}_{(II)}, \\ & \nonumber \underbrace{\big|\hat{T}(s,a,s')-(\hat{T}\circ k)(s,a,s')\big|}_{(III)} \Big\}.
\end{align}

For the moment consider $\hat{T}(s,a,\cdot)$ and $\hat{T}\big(k_{\sigma}(s,a,s'),k_{\alpha}(s,a,s'),\cdot\big)$ just as two sets of numbers. Remove the value corresponding to $s'$ from the first set, the one corresponding to $k_{\sigma'}(s,a,s')$ from the second set, and any remaining zeros from both.
Taking the max between \textit{(I)} and \textit{(II)} just equates to selecting the maximum possible difference between any two values of these modified sets.
Equation \ref{eq:distance} simply tells us to select the worst possible case since we do not know which permutations of states we should compare when computing the Chebyshev distance. $s'$ is removed from $\hat{T}(s,a,\cdot)$ and $k_{\sigma'}(s,a,s')$ is removed from $\hat{T}\big(k_{\sigma}(s,a,s'),k_{\alpha}(s,a,s'),\cdot\big)$ since we know that $k$ maps $(s,a,s')$ to $\big(k_{\sigma}(s,a,s'),k_{\alpha}(s,a,s'),k_{\sigma'}(s,a,s')\big)$ and hence we can compare those values directly \textit{(III)}.

Notice that
\begin{equation}
    \label{eq:bound}
    0 < d_k(s,a,s') \leq 1\text{\hspace{3mm}}\forall (s,a,s') \in S\times A \times S.
\end{equation}
In the following, we propose to improve the algorithms proposed in \cite{icart}. In detail, we redefine the symmetry confidence value $\nu_k$.
We propose to estimate $\nu_k$ as in Line 2 of Algorithm \ref{algo:disc} as:
\begin{equation}
    \label{eq:nu}
    \nu_{k}(\mathcal{D}) = 1-\displaystyle \dfrac{1}{|\mathcal{D}|}\sum_{(s,a,s') \in \mathcal{D}} d_k(s,a,s').
\end{equation}
From equations \ref{eq:distance} and \ref{eq:bound}, it follows that: (i) in deterministic environments $\nu_k$ (Eq. \ref{eq:nu}) coincides with the one prescribed in \cite{icart}; and, (ii) $1 > \nu_k \geq 0$, so $\nu_k$ can be interpreted as a percentage.
This last allows us to suppose that $\nu_k$ is an estimate of the probability of $k$ being a symmetry of the dynamics, and therefore we can relax the necessity of defining an expert-given threshold $\nu$ (cf. \cite{icart} Alg. 1). We then set  $\nu=0.5$ as an input in Algorithm \ref{algo:disc} and eventually augment the batch if $\nu_k > 0.5$ (Lines 3-5).

\begin{remark}[Extreme case scenario]
\textit{Is Equation \ref{eq:distance} too pessimistic?} Consider that for a given state action couple $(\overline{s},\overline{a})$ we have a transition distributed over 3 states $s \in S = \{\textit{One}, \textit{Two}, \textit{Three}\}$ with probabilities $T(\overline{s},\overline{a},One)=0.01$, $T(\overline{s},\overline{a},Two)=0.01$ and $T(\overline{s},\overline{a},Three)=0.98$. Now, assume the estimate of the transition function is perfect. Does the distance in Equation \ref{eq:distance} converge to $0$? Not always, but what matters for the detection of symmetries is the average of the distances over the whole batch (Eq. \ref{eq:nu}).
Suppose that these probabilities were inferred from a batch with the transition $(\overline{s},\overline{a},\textit{One})$ once, $(\overline{s},\overline{a},\textit{Two})$ once and $(\overline{s},\overline{a},\textit{Three})$ ninety-eight times.
 Consider $(\overline{s},\overline{a},\textit{Three})$. $M(\overline{s},\overline{a},\textit{Three}) = M_k(\overline{s},\overline{a},\textit{Three}) = m(\overline{s},\overline{a},\textit{Three}) = m_k(\overline{s},\overline{a},\textit{Three})=0.01$.
 Following Eq. \ref{eq:distance}, $d_k(\overline{s},\overline{a},\textit{Three})=0$.
 However, $d_k(\overline{s},\overline{a},\textit{One})=d_k(\overline{s},\overline{a},\textit{Two})=0.97$, which is a too pessimistic estimate. 
Nevertheless let's calculate $\nu_k$ (Eq.\ref{eq:nu}).
For this state-action pair $(\overline{s},\overline{a})$, the average over the batch is therefore: 
$( d_k(\overline{s},\overline{a},\textit{One}) + d_k(\overline{s},\overline{a},\textit{Two}) + 98 d_k(\overline{s},\overline{a},\textit{Three}) )/100 = 0.0194$.
If the estimation is the same for other pairs $(s,a)$, then $\nu_k=1-0.0194=0.9806$.
This is a value close to 1 suggesting $k$ is a symmetry.
\end{remark}
\input{algorithms/discrete}
\input{algorithms/continuous}

%% file: algorithms/discrete.tex
\begin{algorithm}[!t]
\SetAlgoLined
\KwInput{Batch of transitions $\mathcal{D}$, $k$ alleged symmetry}
\KwOutput{Possibly augmented batch $\mathcal{D}\cup \mathcal{D}_k$}
 $\hat{T} \leftarrow$ Most Likely Categorical pmf from $\mathcal{D}$\\
$\nu_{k} = 1-\displaystyle \dfrac{1}{|\mathcal{D}|}\sum_{(s,a,s') \in \mathcal{D}} d_k(s,a,s')$ \hspace{0.5cm} \textit{(where $d_k$ is defined in Equation \ref{eq:distance})}\\
 \eIf{$\nu_k > 0.5$}{
    $\mathcal{D}_k = k(\mathcal{D})$ \hspace{0.02cm} \textit{(alleged symm. transitions)}\\
  \textbf{return}
  $\mathcal{D}\cup \mathcal{D}_k$ \hspace{0.5cm} \textit{(the augmented batch)}}
   {\textbf{return} $\mathcal{D}$ \hspace{0.5cm} \textit{(the original batch)}
  }

 \caption{Symmetry detection and data augmenting in a categorical MDP}
 \label{algo:disc}
\end{algorithm}

%% file: algorithms/continuous.tex
\begin{algorithm}
\SetAlgoLined
\KwInput{Batch of transitions $\mathcal{D}$, $q \in [0,1)$ order of the quantile, $k$ alleged symmetry}
\KwOutput{Possibly augmented batch $\mathcal{D}\cup \mathcal{D}_k$}
 $\mathcal{L} \leftarrow$ Density Estimate ($\mathcal{D}$) \hspace{0.1cm} \textit{(e.g. with Normalizing Flows)}\\
 $\Lambda \leftarrow$ Distribution $\mathcal{L}(\mathcal{D})$ \hspace{0.02cm}\textit{($\mathcal{L}$ evaluated over $\mathcal{D}$)}\\
 $\theta = q$-order quantile of $\Lambda$\\
 $\mathcal{D}_k = k(\mathcal{D})$ \hspace{0.02cm} \textit{(alleged symmetric transitions)}\\
 $\nu_{k} = \displaystyle \dfrac{1}{|\mathcal{D}_k|}\sum_{(s,a,s') \in \mathcal{D}_k}\mathds{1}_{\{\mathcal{L}(s,a,s'|\mathcal{D})>\theta\}}$\\
 \eIf{$\nu_k > 0.5$}{
  \textbf{return}
  $\mathcal{D}\cup \mathcal{D}_k$ \hspace{0.5cm} \textit{(the augmented batch)}}
   {\textbf{return} $\mathcal{D}$ \hspace{0.5cm} \textit{(the original batch)}
  }

 \caption{Symmetry detection and data augmenting in a continuous MDP with detection threshold $\nu=0.5$ \cite{icart}}
 \label{algo:cont}
\end{algorithm}

%% file: sections/3-experiments.tex
\section{Experiments}
In order to show the improvements provided by our contribution we tested the algorithms in a stochastic version of the toroidal Grid environment  and two continuous state environments of the OpenAI's Gym Learning Suite: CartPole and Acrobot. 
We have chosen the same scenarios as \cite{icart} in order to demonstrate that our approach generalizes well to the stochastic case contrary to the approach proposed in \cite{icart}.

\subsection{Setup}
We collect a batch of transitions $\mathcal{D}$ using a uniform random policy. An expert alleges the presence of a symmetry $k$ and we proceed to its detection using Algorithm \ref{algo:disc} (categorical case) or Algorithm \ref{algo:cont} (continuous case).
In the continuous case, Density Estimation is performed by a Masked Autoregressive Flow architecture \cite{maf} with $3$ layers of bijectors.

The experiments were performed using 2 Dodeca-core Skylake Intel\textsuperscript\tiny{\textregistered}\normalsize\  Xeon\textsuperscript\tiny{\textregistered}\normalsize\  Gold 6126 @ 2.6 GHz and 96 GB of RAM and 2 GPU NVIDIA\textsuperscript\tiny{\textregistered}\normalsize\ V100 @ 192GB of RAM. The code to run the experiments is available at \href{https://github.com/giorgioangel/dsym}{https://github.com/giorgioangel/dsym}.

\paragraph{Computation of $\nu_k$ and batch augmentation}
We report the ${\nu}_k$ obtained with an ensemble of $N$ different iterations of the procedure: we generate $z\in \mathbb{N}$ sets of $N$ different batches $\mathcal{D}$ of increasing size.
Remember that since $\nu_k \in [0,1)$ we can interpret it as the probability of the presence of a symmetry and select a detection threshold $\nu = 0.5$ or higher, while in \cite{icart} the threshold $\nu$ was expert-given. We calculate $\nu_{k}$ with both the \cite{icart} method and the approach here presented.

\paragraph{Evaluation of the performance (Categorical case)}
In the end, let $\rho$ be the distribution of initial states $s_0 \in S$ and let the performance $U^{\pi}$ of a policy $\pi$ be
    $U^{\pi} = \mathbb{E}_{s\sim \rho}[V^{\pi}(s)].$
Our experimental contribution is the comparison between the performances obtained by acting in the real environment with $\hat{\pi}$ (the optimal policy solving the MDP defined with $\hat{T}$) and $\hat{\pi}_k$ (the optimal policy obtained with $\hat{T}_k$). In particular we consider the quantity
\begin{equation}
\label{eq:deltaperf}
\Delta U = U^{\hat{\pi}_{k}} - U^{\hat{\pi}}.
\end{equation}
$\Delta U > 0$ means that data augmenting leads to better policies.

In \textit{categorical} environments the policies are obtained with Policy Iteration and evaluated with Policy Evaluation.
\paragraph{Evaluation of the performance (Continuous case)}
In \textit{continuous} environments Offline Learning is not trivial. We use the implementation of two Model-Free Deep RL architectures: Deep Q-Network (DQN) \cite{mnih2015human} and Conservative Q-Learning (CQL) \cite{kumar2020conservative} of the d3rlpy learning suite \cite{seno2021d3rlpy} to obtain a policy starting from the batches. The first method is the one that originally established the validity of Deep RL and it is used in online RL while the second was specifically developed to tackle offline RL problems. Since the convergence of the training of Deep RL baselines is greatly dependent of hyperparameter tuning that itself depends on both the environment and the batch \cite{paine2020hyperparameter}, we will apply DQN and CQL with the default parameters provided by d3rlpy, abiding hence more faithfully to an offline learning duty. This means that sometimes the learning might not converge to a good policy. We find this philosophy more honest than showing the results obtained with the best seed or the finest-tuned hyperparameters. Each architecture is trained for a number of steps equal to fifty times the number of transitions present in the batch.
\subsection{Environments}
\paragraph{Stochastic Grid (Categorical)}
In this environment, the agent can move along fixed directions over a torus by acting with any $ a\in A = \{\uparrow, \downarrow, \leftarrow, \rightarrow\}$ (see Figure \ref{fig:torus}). The grid meshing the torus has size $l=10$. \input{figures/torus.tex}
The agent can spawn everywhere on the torus with a uniform probability and must reach a fixed goal. At every time step, the agent receives a reward $r=-1$ if it does not reach the goal and a reward $r=1$ once the goal has been reached, terminating the episode. When performing an action the agent has $60\%$ chances of moving to the intended direction, $20\%$ to the opposite one, and $10\%$ along an orthogonal direction. We collect $z=10$ sets of $M=100$ batches with respectively $N = 1000 \times i_z $ steps in each batch ($i_z$ going from $1$ to $z$).
\input{tables/grid-transformations}

\begin{figure}[!h]
    \centering
    \includegraphics[bb=0 0 469.44 307.44, width=\columnwidth]{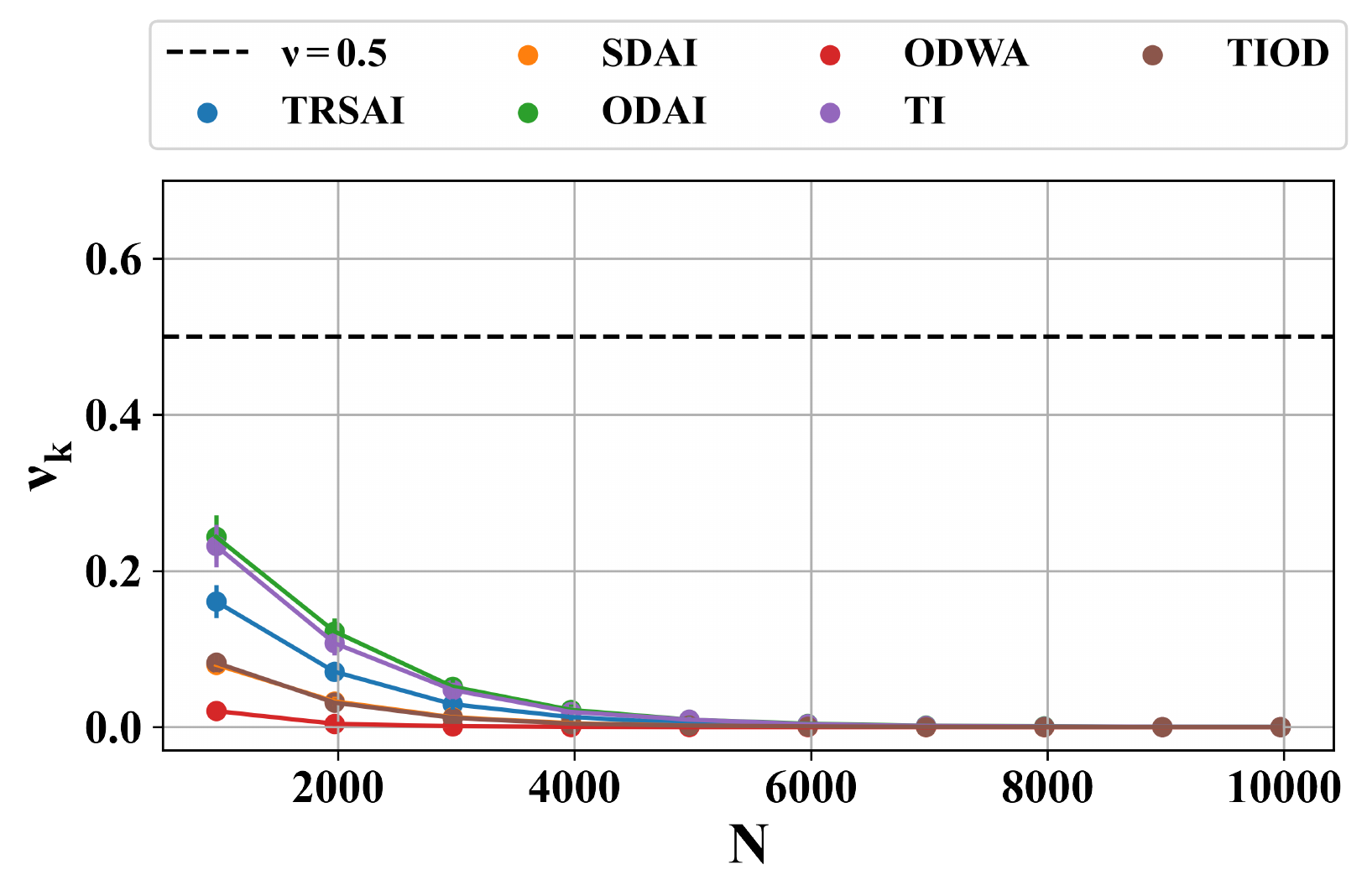}
    \caption{Stochastic toroidal Grid Environment. Probability of symmetry $\nu_k$ calculated with the method proposed by \cite{icart}. The threshold at $\nu=0.5$ is displayed as a dashed line. Since all $\nu_k<0.5$ means that no transformation is detected as a symmetry.}
\label{fig:nu_sota}
\end{figure}

The proposed symmetries for this environment are outlined in Table \ref{tab:grid-transf}. We check for the invariant of the dynamics with respect to the following six transformations (the valid symmetries are displayed in \textbf{bold}): (1) Time reversal symmetry with action inversion \textbf{(TRSAI)}; (2) Same dynamics with action inversion (SDAI); (3) Opposite dynamics and action inversion \textbf{(ODAI)}; (4) Opposite dynamics but wrong action (ODWA); (5) Translation invariance \textbf{(TI)}; (6) Translation invariance with opposite dynamics (TIOD).
The $N$ dependent average results for symmetry detection using the method from \cite{icart} are reported in Figure \ref{fig:nu_sota}, and results using our method are displayed in Figure  \ref{fig:res_grid_nu}. Figure \ref{fig:res_grid_perf} presents the performance improvement $\Delta U$, with its standard deviation being represented by a vertical error bar.
\input{figures/grid_results}
\input{figures/ab-cp-nu}

\paragraph{Stochastic CartPole (Continuous)}
A pole is precariously balanced on a cart and an agent can push the whole system left or right to prevent the pole from falling. \input{figures/cartpole.tex}
The dynamics is similar to that of CartPole \cite{brockman2016openai} (see Figure \ref{fig:cartpole}), however the force that the agent uses to push the cart is sampled from a normal distribution with mean $f$ (the force defined in the deterministic version) and standard deviation $\tilde{\sigma} = 2$. Recall that the state is represented by the features $(x, \theta, v, \omega)$ and $A=\{\leftarrow, \rightarrow\}$. For the evaluation of $\nu_k$ we set the quantile $q=0.1$ and we collect $z=10$ sets of $M=100$ batches with respectively $N = 1000 \times i_z $ steps in each batch (and $i_z$ going from $1$ to $10$). We evaluate $\Delta U$ by training the agent on single batches of $N = 5000 \times i_z $ (and $i_z$ going from $1$ to $6$) both augmented and not augmented with $k$.

The acronyms of the valid symmetric transformations are displayed in \textbf{bold}: (1) State and action reflection with respect to an axis in $x=0$ \textbf{(SAR)}; (2) Initial state reflection (ISR); (3) Action inversion (AI); (4) Single feature inversion (SFI); (5) Translation invariance \textbf{(TI)}.
Their effects on the transition $(s,a,s')$ are listed in Table \ref{tab:cp-tra}. Average results and errors are displayed in Figure \ref{fig:res_cp_nu}. The results considering the evaluation of performance gain ($\Delta U$) are
shown in Table \ref{tab:cartpole-perf}.

\paragraph{Stochastic Acrobot (Continuous)}
The Acrobot is a planar two-link robotic arm working against gravity, the agent can decide whether to swing or not the elbow left or right to balance the arm straightened up (see Figure \ref{fig:acrobot}). It is the very same Acrobot of \cite{brockman2016openai} but at every time step a noise $\epsilon$ is sampled from a uniform distribution on the interval $[-0.5, 0.5]$ and added to the torque. A state is represented by the features $(s_1, c_1, s_2, c_2, \omega_1, \omega_2)$ where $s_i$ and $c_i$ are respectively $sin(\alpha_i)$ and $cos(\alpha_i)$ in shorthand notation. The action set $A=\{-1, 0, 1\}$. For the evaluation of $\nu_k$ we set $q=0.1$. For the detection case, we collected $z=5$ sets of $M=100$ batches with $N=1000 \times i_z$ steps within each one ($i_z$ going from $1$ to $z$). The evaluation of the performance was carried out on single batches, with and without data augmentation, with $N=10000 \times i_z$ steps and $i_z$ going from $1$ to $4$. For the evaluation of $\Delta$ $z=5$ due to computational necessities.
We allege the following transformations $k$, as always the valid ones are \textbf{bolded}: (1) Angles and angular velocities inversion \textbf{(AAVI)}; (2) Cosines and angular velocities inversion (CAVI); (3) Action inversion (AI); (4) Starting state inversion (SSI).
\input{figures/acrobot.tex}
The images of the transformations are reported in Table \ref{tab:ab-tra}. The $N$ dependent average results and standard deviations are reported in Figure \ref{fig:res_ab_nu}. The results considering the evaluation of performance gain ($\Delta U$) are
shown in Table \ref{tab:acrobot-perf}.

\input{tables/cp-ab}

%% file: figures/torus.tex
\begin{figure}[bp!]
\centering
\begin{tikzpicture}[scale=0.8, transform shape]
    \begin{axis}[hide axis]
       \addplot3[surf,
       colormap/blackwhite,
       samples=20,
       domain=0:2*pi,y domain=0:2*pi,
       z buffer=sort]
       ({(2+cos(deg(x)))*cos(deg(y+pi/2))}, 
        {(2+cos(deg(x)))*sin(deg(y+pi/2))}, 
        {sin(deg(x))});
        \addplot3[color=red, thick, ->] coordinates {(-0.85,-0.85,0) (-0.85,-0.85,0.25)} ;
        \addplot3[mark=*,red,point meta=explicit symbolic,nodes near coords] 
coordinates {(-0.85,-0.85,0)[]};
    \end{axis}
\end{tikzpicture}
\caption{Representation of the Grid Environment \cite{icart}. The red dot is the position of a state $s$ on the torus. A possible displacement obtained by acting with action $a=\uparrow$ is shown as a red arrow.}\label{fig:torus}
\end{figure}

%% file: tables/grid-transformations.tex
\normalsize
\begin{table}[th]
\caption{Toroidal Grid: proposed transformations and label.}\label{tab:grid-transf} \centering
\resizebox{0.95\columnwidth}{!}{%
\begin{tabularx}{1.1\columnwidth}{lY}
  \hline
  \noalign{\vskip 0.1mm}$k$ & Label\\
  \hline
  \hline
  \small$k_{\sigma}(s,a,s') = s'$ & \\
  \small$ k_{\alpha}\big(s,a=(\uparrow, \downarrow, \leftarrow, \rightarrow),s'\big)  = (\downarrow, \uparrow, \rightarrow, \leftarrow)$ & \textbf{TRSAI}\\\small$k_{\sigma'}(s,a,s') = s$  & \\
  \hline
  \small$k_{\sigma}(s,a,s') = s$ & \\\small$ k_{\alpha}\big(s,a=(\uparrow, \downarrow, \leftarrow, \rightarrow), s' \big)  = (\downarrow, \uparrow, \rightarrow, \leftarrow)$ & SDAI\\\small$k_{\sigma'}(s,a,s') = s'$  & \\
  \hline
  \small$k_{\sigma}(s,a,s') = s$ & \\\small$ k_{\alpha}\big(s,a=(\uparrow, \downarrow, \leftarrow, \rightarrow),s' \big)  = (\downarrow, \uparrow, \rightarrow, \leftarrow)$ & \textbf{ODAI}\\\small$k_{\sigma'}(s,a=(\uparrow, \downarrow, \leftarrow, \rightarrow),s') =$ & \\
  \ \ $\big(s'-(0,2),s'+(0,2),s'+(2,0),s'-(2,0) \big)$ & \\
  \hline
  \small$k_{\sigma}(s,a,s') = s$ & \\\small$ k_{\alpha}(s,a=(\uparrow, \downarrow, \leftarrow, \rightarrow),s')  = (\rightarrow, \leftarrow, \uparrow, \downarrow)$ & ODWA\\\small$k_{\sigma'}(s,a=(\uparrow, \downarrow, \leftarrow, \rightarrow),s') =$ & \\
  \ \ $\big(s'-(0,2),s'+(0,2),s'+(2,0),s'-(2,0) \big)$ & \\
  \hline
  \small$k_{\sigma}(s,a,s') = s'$ & \\\small$ k_{\alpha}(s,a,s')  = a$ & \textbf{TI}\\\small$k_{\sigma'}\big(s,a=(\uparrow, \downarrow, \leftarrow, \rightarrow),s'\big) =$ & \\ \ \ $\big(s'+(0,1), s'-(0,1), s'-(1,0), s'+(1,0) \big)$  & \\
  \hline
  \small$k_{\sigma}(s,a,s') = s'$ & \\\small$ k_{\alpha}(s,a,s')= a $ & TIOD\\\small$k_{\sigma'}(s,a,s') = s$  & \\
  \hline
\end{tabularx}%
}
\end{table}
\normalsize

%% file: figures/grid_results.tex
\begin{figure}[t!]
\centering
    \begin{subfigure}[t]{0.47\textwidth}
         \centering
         \includegraphics[bb=0 0 469.44 307.44, width=\textwidth]{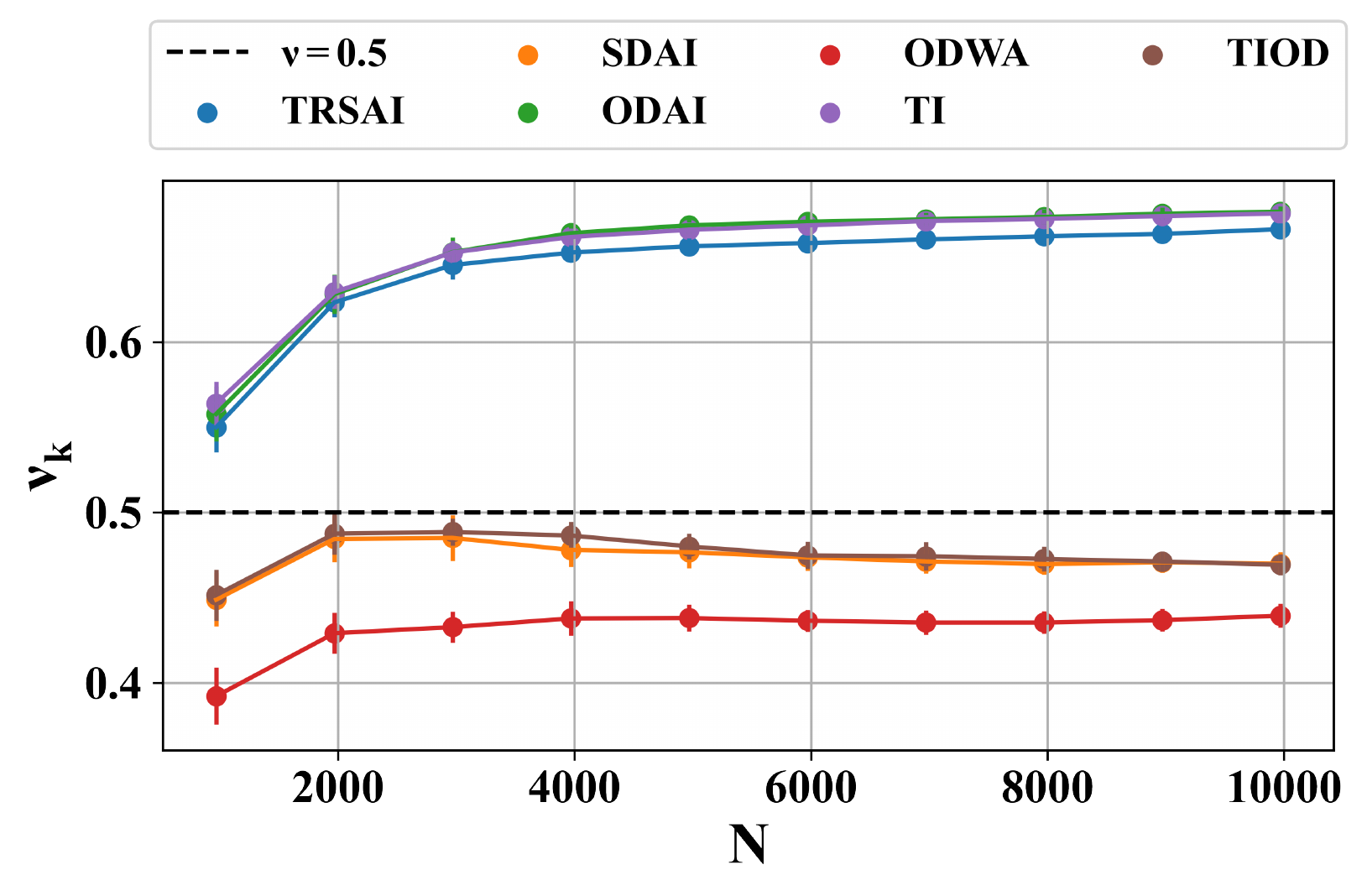}
         \caption{Probability of symmetry $\nu_k$ \textbf{with our approach}. The threshold at $\nu=0.5$ is displayed as a dashed line. $\nu_k>0.5$ means that the transformation is detected as a symmetry.}
         \label{fig:res_grid_nu}
     \vspace{4mm}
     \end{subfigure}
     
     \begin{subfigure}[t]{0.47\textwidth}
         \centering
         \includegraphics[bb=0 0 469.44 307.44, width=\textwidth]{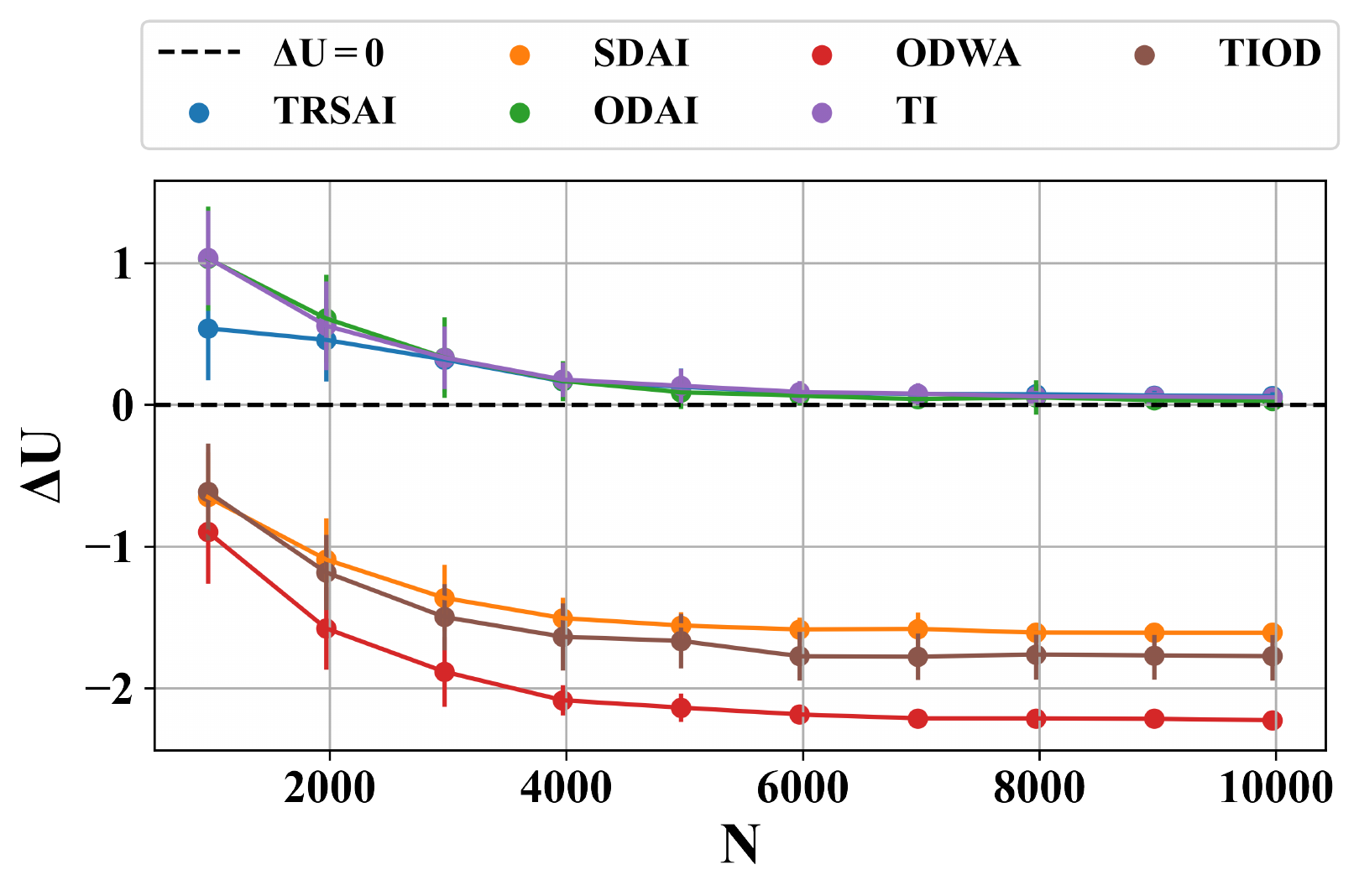}
         \caption{Performance difference $\Delta U$ (Eq. \ref{eq:deltaperf}). The threshold at $\Delta U=0$ is displayed as a dashed line. $\Delta U>0$ means that data augmenting leads to better policies.}
         \label{fig:res_grid_perf}
     \end{subfigure}
     \vspace{3mm}
\caption{Stochastic Toroidal Grid Environment. $\nu_k$ and $\Delta U$ for the transformations $k$ computed over sets of $100$ different batches of size $N$. Points are mean values and bars standard deviations.}
\label{fig:res_grid}
\end{figure}

%% file: figures/ab-cp-nu.tex
\begin{figure}[t!]
\centering
    \begin{subfigure}{0.47\textwidth}
         \centering
         \includegraphics[bb=0 0 469.44 307.44, width=\textwidth]{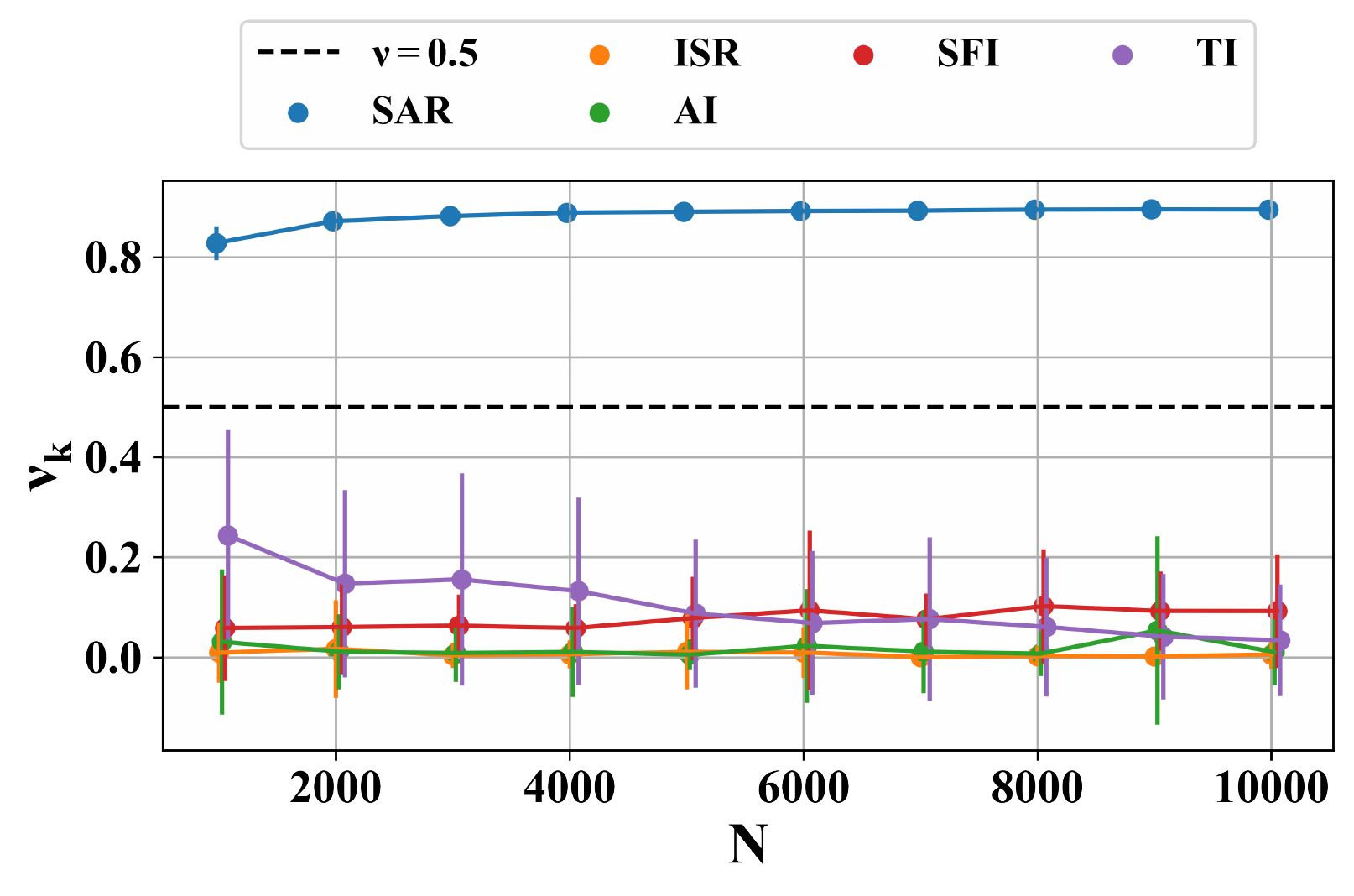}
         \caption{Stochastic CartPole. Probability of symmetry $\nu_k$. The threshold at $\nu=0.5$ is displayed as a dashed line. $\nu_k>0.5$ means that the transformation is detected as a symmetry.}
         \label{fig:res_cp_nu}
     \end{subfigure}%
     \vskip 5pt
    \begin{subfigure}{0.47\textwidth}
         \centering
         \includegraphics[bb=0 0 469.44 307.44, width=\textwidth]{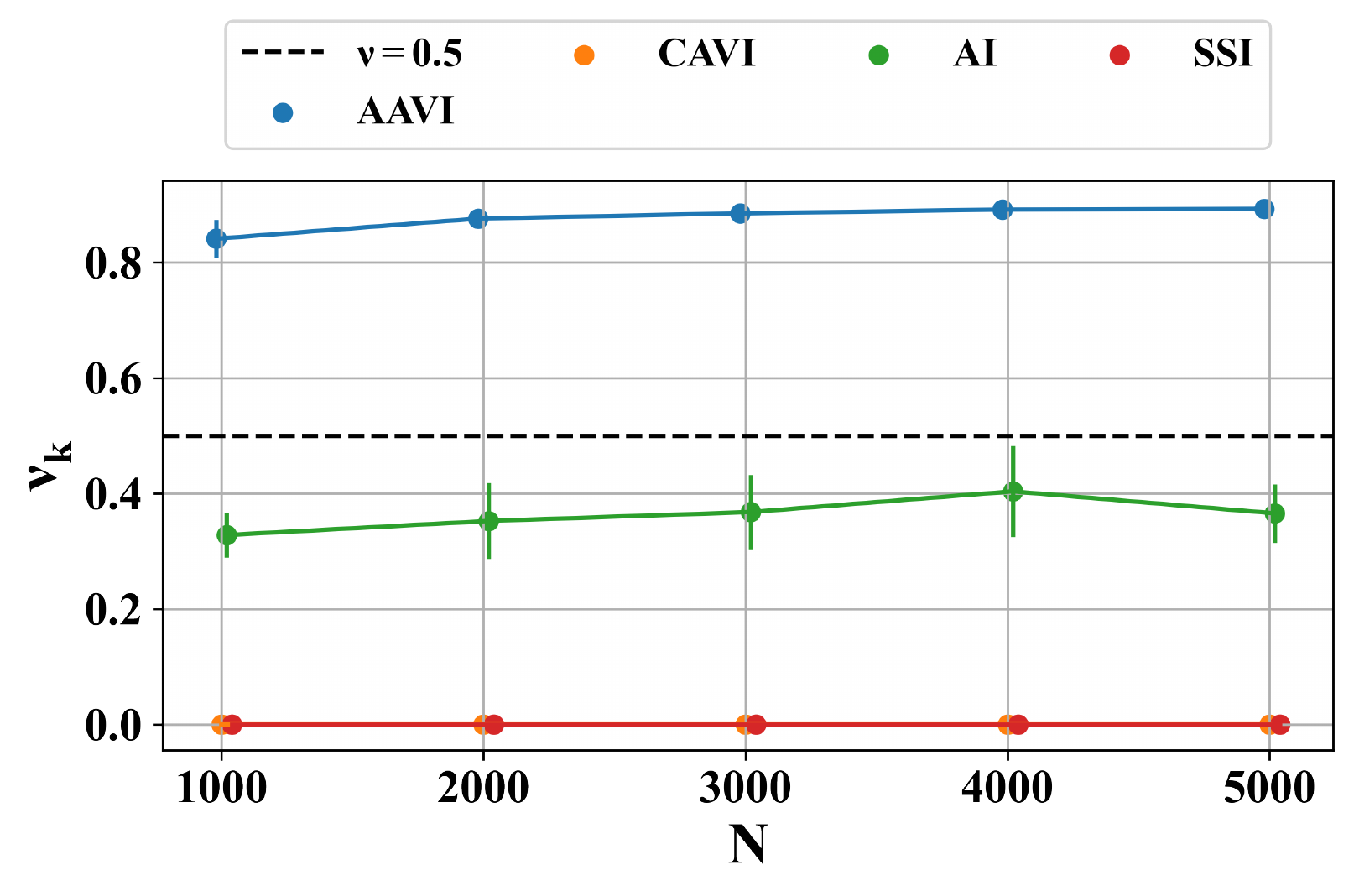}
         \caption{Stochastic Acrobot. Probability of symmetry $\nu_k$. The threshold at $\nu=0.5$ is displayed as a dashed line. $\nu_k>0$ means that the transformation is detected as a symmetry.}
         \label{fig:res_ab_nu}
     \end{subfigure}
     \vskip 10pt
\caption{$\nu_k$, for the transformations $k$ computed over sets of different batches of size $N$ in Stochastic CartPole (left) and Stochastic Acrobot (right). Points are mean values and are a bit shifted horizontally for the sake of display. Standard deviation is displayed as a vertical error bar.}
\label{fig:res_nu}
\end{figure}

%% file: figures/cartpole.tex
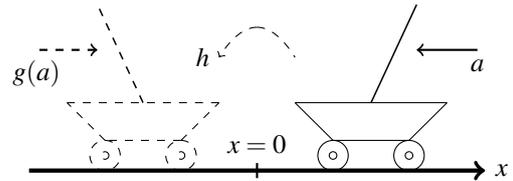
\begin{figure}
\centering
    \begin{tikzpicture}
\draw[->,ultra thick] (-3,0)--(3,0) node[right]{$x$};
\draw [thick] (0, -0.1) -- (0, 0.1) node[above]{$x=0$};
\draw (1, 0.2) circle (0.2cm);
\draw (2, 0.2) circle (0.2cm);

\draw (1, 0.2) circle (0.05cm);
\draw (2, 0.2) circle (0.05cm);

\draw (1, 0.4) -- (0.5, 0.9);
\draw (2, 0.4) -- (2.5, 0.9);
\draw (1, 0.4) -- (2, 0.4);
\draw (0.5, 0.9) -- (2.5, 0.9);

\draw [line width=0.02 cm] (1.5, 0.9) -- (2.1, 2.2);

\draw [<-, thick] (2.1, 1.6) -- (2.9, 1.6) node[below]{$a$};

\draw [dashed] (0, 1.9) parabola (0.5, 1.5);
\draw [->, dashed] (0, 1.9) parabola (-0.5, 1.5) node[above, left]{$h$};

\draw [dashed] (-1, 0.2) circle (0.2cm);
\draw [dashed] (-2, 0.2) circle (0.2cm);

\draw [dashed] (-1, 0.2) circle (0.05cm);
\draw [dashed] (-2, 0.2) circle (0.05cm);

\draw [dashed] (-1, 0.4) -- (-0.5, 0.9);
\draw [dashed] (-2, 0.4) -- (-2.5, 0.9);
\draw [dashed] (-1, 0.4) -- (-2, 0.4);
\draw [dashed] (-0.5, 0.9) -- (-2.5, 0.9);
\draw [line width=0.02 cm, dashed] (-1.5, 0.9) -- (-2.1, 2.2);

\draw [<-, thick, dashed] (-2.1, 1.6) -- (-2.9, 1.6) node[below]{$g(a)$};
    \end{tikzpicture}
    \caption{The cart in the right is a representation of a CartPole's state $s_t$ with $x_t > 0$ and action $a_t=\leftarrow$ \cite{icart}. The dashed cart in the left is the image of $(s_t,a_t)$ under the transformation $h$ which inverses state $f(s)=-s$ and action $g(a)=-a$.}\label{fig:cartpole}
\end{figure}

%% file: figures/acrobot.tex
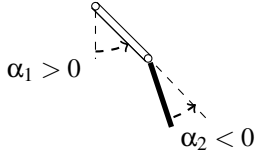
\begin{figure}
\centering
    \begin{tikzpicture}
    \draw[dashed] (0.05, -0.08) -- (-0.64, 0.65);
    \draw[dashed] (-1.4, 1.35) -- (-1.4, 0.7);
    \draw[rotate=45] (-0.05, 1) -- (-0.05, 2);
    \draw[rotate=45] (0.05, 1) -- (0.05, 2);
    \draw[rotate=45] (-0.05, 1) -- (-0.05, 2);
    \draw[line width=0.7mm] (-0.68, 0.65) -- (-0.4, -0.2) ;
    \draw[rotate=45] (0, 1) circle (0.05);
    \draw (-1.4, 1.4) circle (0.05);
    \draw[thick, dashed, ->] (-1.4, 0.8) arc (270:296:1) node[below=0.35cm, left=0.5cm]{$\alpha_1>0$};
    \draw[thick, dashed, <-] (-0.08, 0.1) arc (310:288:1) node[below right]{$\alpha_2<0$};
    \end{tikzpicture}
    \caption{Representation of a state of the Acrobot environment \cite{icart}.}\label{fig:acrobot}
\end{figure}

%% file: tables/cp-ab.tex
\begin{table}
\centering
\caption{Proposed transformations and labels for Stochastic CartPole. \label{tab:cp-tra}}
\begin{minipage}{\columnwidth}
\begin{tabularx}{\columnwidth}{lY}
  \hline
  \noalign{\vskip 0.1mm}$k$ & Label\\
    \hline
    \hline
  \small$k_{\sigma}(s,a,s') = -s$ &   \\\small$k_{\alpha}\big(s,a=(\leftarrow, \rightarrow),s'\big) = (\rightarrow, \leftarrow)$ & \textbf{SAR} \\\small$k_{\sigma'}(s,a,s') = -s'$ &  \\
    \hline
  \small$k_{\sigma}(s,a,s') = -s$ &  \\\small$k_{\alpha}(s,a,s') = a$ & ISR \\\small$k_{\sigma'}(s,a,s') = s'$  & \\
    \hline
  \small$k_{\sigma}(s,a,s') = s$ & \\\small$k_{\alpha}\big(s,a=(\leftarrow, \rightarrow), s'\big)= (\rightarrow, \leftarrow)$ & AI \\\small$k_{\sigma'}(s,a,s') = s'$  & \\
    \hline
  \small$k_{\sigma}\big(s=(x, ...),a,s'\big) = (-x, ...)$ & \\\small$k_{\alpha}(s,a,s')=a$ & SFI\\\small$k_{\sigma'}(s,a,s') = s'$  & \\
      \hline
    \small$k_{\sigma}\big( s =(x, ...), a, s' \big) = (x+0.3, ...)$ & \\\small$k_{\alpha}(s,a,s') = a$ & \textbf{TI}\\\small$k_{\sigma'}\big(s,a,s'=(x', ...) \big) = (x'+0.3, ...)$
\end{tabularx}
\end{minipage}
\end{table}
\begin{table}
\caption{Proposed transformations and labels for Stochastic Acrobot. \label{tab:ab-tra}}
\begin{minipage}{\columnwidth}
\begin{tabularx}{\columnwidth}{l Y}
  \hline
  \noalign{\vskip 0.1mm}$k$ & Label\\
    \hline
    \hline
  \small$k_{\sigma}\big(s=(s_1, s_2, \omega_1, \omega_2, \dots),a,s'\big)$ & \\
  \hspace{1cm} $=(-s_1, -s_2, -\omega_1, -\omega_2, \dots)$ &
 \\\small$k_{\alpha}(s,a=(-1, 0, 1),s'\big) = (1, 0, -1)$ & \textbf{AAVI} \\\small$k_{\sigma'}\big(s,a,s'=(s'_1, s'_2, \omega'_1, \omega'_2, \dots)\big) $  & \\
 \hspace{1cm} $= (-s'_1, -s'_2, -\omega'_1, -\omega'_2, \dots)$ & \\
    \hline
    \small$k_{\sigma}\big(s=(c_1, c_2, \omega_1, \omega_2, \dots),a,s'\big)$ & \\ 
    \hspace{1cm} $= (-c_1, -c_2, -\omega_1, -\omega_2, \dots)$ & \\
    \small$k_{\alpha}\big(s,a=(-1, 0, 1),s'\big) = (1, 0, -1)$ & CAVI \\\small$k_{\sigma'}\big(s,a,s'=(c'_1, c'_2, \omega'_1, \omega'_2, \dots)\big)$  & \\
    \hspace{1cm} $= (-c'_1, -c'_2, -\omega'_1, -\omega'_2, \dots)$ & \\
    \hline
    \small$k_{\sigma}(s,a,s') = s$ & \\\small$k_{\alpha}\big(s,a=(-1, 0, 1),s'\big) = (1, 0, -1)$ & AI\\\small$k_{\sigma'}(s,a,s') = s'$  & \\
    \hline
      \small$k_{\sigma}(s,a,s') = -s$ & \\\small$k_{\alpha}\big(s,a,s') = a$ & SSI\\\small$k_{\sigma'}(s,a,s') = s'$  & \\
        \hline

\end{tabularx}
   \end{minipage}
\end{table}

%% file: sections/4-discussion.tex
\section{Discussion}
\paragraph{Stochastic Grid (Categorical)} \textit{Detection phase ($\nu_k$).}
We see from Figure \ref{fig:nu_sota} that using the state-of-the-art approach no transformation is detected as a symmetry because $\nu_k < 0.5$, $\forall k$ in the proposed set of transformations. This result highlights the inadequacy of the state-of-the-art method to deal with stochastic environments. On the contrary, our novel algorithm perfectly manages to identify the real symmetries of the environment (see Figure \ref{fig:res_grid_nu}): $\nu_k > 0.5$, $k\in\{\textrm{TRSAI}, \textrm{ODAI}, \textrm{TI}\}$. Moreover, there are no false positives: $\nu_k < 0.5$,  $k\in\{\textrm{SDAI}, \textrm{ODWA}, \textrm{TIOD}\}$. We notice that while in a deterministic environment $\nu_k = 0$ $\forall k$ which is not a symmetry, here the stochasticity makes the detection more complicated since $\nu_k \approx 0.5^{-}$ for $N=2000$.

\input{tables/cartpole-perf}

\textit{Evaluation of performance gain ($\Delta U$).}
The difference in the performance of the deployed policies $\Delta U$ perfectly fits the expected behavior. When $k$ is a symmetry $\Delta U>0$ and saturates to $0$ with $N$ increasing. When $k$ is not a symmetric transformation of the dynamics $\Delta U <0$ and keeps decreasing with $N$ (see Figure \ref{fig:res_grid_perf}).

\paragraph{Stochastic CartPole (Continuous)} \textit{Detection phase ($\nu_k$)} In Stochastic CartPole the algorithm fails to detect the symmetry $k=\textrm{TI}$. This could be due to the fact that the translation invariance symmetry in this case is fixed for a specific value (see TI in Table \ref{tab:cp-tra} where the translation is set at $0.3$). If the translation is too small the neural network fails to discern the transformation from the noise.
The algorithm classifies correctly as a symmetry $k=\textrm{SAR}$ and the remaining transformations as non-symmetries (see Figure \ref{fig:res_cp_nu}).

\textit{Evaluation of performance gain ($\Delta U$).}
Results are displayed in Table \ref{tab:cartpole-perf}. ORL is very unstable and sensitive to the choice of hyperparameters. On top of that, the training is carried out for a fixed number of epochs. We notice that, on average over different batch sizes, $\Delta U > 0$ for DQN and  SAR, and SFI transformations. While SAR is a valid symmetry, SFI it's not. A more conservative algorithm like CQL only detects SAR as a valid symmetry. The performance difference for TI both for DQN and CQL is so close to zero that we think that augmenting the dataset with this symmetry might not be a substantial power-up over using just the information contained in the original batch.

\paragraph{Stochastic Acrobot (Continuous)} \textit{Detection phase ($\nu_k$).} In this environment the only real symmetry of the dynamics, $\textit{AAVI}$, gets successfully detected by the algorithm with $q=0.1$. Non symmetries yield a $\nu_k<0.5$ (Figure \ref{fig:res_ab_nu}).

\input{tables/acrobot-perf}
\textit{Evaluation of performance gain ($\Delta U$).}
Results are displayed in Table \ref{tab:acrobot-perf} and show that the training in Stochastic Acrobot is harder than in Stochastic CartPole since, even with a large dataset, sometimes the algorithms do not manage to learn a good policy. In particular, while CQL manages to learn how to behave in the environment exploiting the \textbf{AAVI} symmetry (average $\Delta U = 52.9$), DQN still struggles with every $k$, good and wrong. Nevertheless, CQL apparently benefits from augmenting the dataset also with wrong symmetries even though to a smaller extent. We suppose this effect is due to the instability in ORL training.

%% file: tables/cartpole-perf.tex
\begin{table*}[h]
\centering
\caption{$\Delta U$ for every alleged symmetry in Stochastic CartPole with two baselines and different batch sizes $N$.}
\label{tab:cartpole-perf}
\begin{tabular}{@{}cclllllll@{}}
\toprule
\multicolumn{1}{l}{}         &               & \multicolumn{6}{c}{\textit{$N$ (number of transitions in the original batch)}}                     &                       \\ \cmidrule(l){3-8} 
\textit{k} &
  \textit{Baseline} &
  \multicolumn{1}{c}{5000} &
  \multicolumn{1}{c}{10000} &
  \multicolumn{1}{c}{15000} &
  \multicolumn{1}{c}{20000} &
  \multicolumn{1}{c}{25000} &
  \multicolumn{1}{c}{30000} &
  \multicolumn{1}{c}{\textit{Average $\Delta U$}} \\ \midrule
\multirow{2}{*}{\textbf{SAR}} & \textit{DQN} & -7.3 & 25.4 & 41.8 & 7.2 & 9.0 & 3.4 & \textbf{13.3}\\
 & \textit{CQL} & 37.4 & -2.5 & -4.1 & 20.1 & 17.9 & -9.0 & \textbf{10.0} \\ \\
\multirow{2}{*}{ISR} & \textit{DQN} & -1.3 & -48.5 & -29.9 & -78.7 & -107.8 & -29.1 & -49.2\\
 & \textit{CQL} & 6.4 & 1.6 & -2.2 & -22.3 & -10.3 & -25.9 & -8.8\\ \\
\multirow{2}{*}{AI} & \textit{DQN} & 26.9 & -48.5 & -43.7 & -74.6 & -41.3 & -84.6 & -44.3\\
 & \textit{CQL} & -13.1 & -7.6 & -29.8 & -6.5 & -22.3 & -15.3 & -15.8\\
 \\
\multirow{2}{*}{SFI} & \textit{DQN} & -33.4 & 17.9 & 21.4 & 45.4 & -6.9 & -0.1 & 7.4\\
 & \textit{CQL} & -5.5 & -2.1 & 7.4 & -3.9 & -3.6 & -18.5 & -4.4\\ \\
\multirow{2}{*}{\textbf{TI}} & \textit{DQN} & 36.9 & -28.1 & 34.5 & 15.7 & 6.1 & -9.1 & -0.2\\
 & \textit{CQL} & 7.6 & -1.3 & -2.1 & 11.8 & -16.5 & 5.2 & \textbf{0.8} \\ \bottomrule
\end{tabular}
\end{table*}

%% file: tables/acrobot-perf.tex
\begin{table*}[]
\vskip 15pt
\centering
\caption{$\Delta U$ for every alleged symmetry in Stochastic Acrobot with two baselines and different batch sizes $N$.\label{tab:acrobot-perf}}
\begin{tabular}{@{}cclllll@{}}
\toprule
\multicolumn{1}{l}{}         &               & %
\multicolumn{4}{c}{\textit{$N$}} &  \\ \cmidrule(l){3-6} 
\textit{k} &
  \textit{Baseline} &
  \multicolumn{1}{c}{10000} &
  \multicolumn{1}{c}{20000} &
  \multicolumn{1}{c}{30000} &
  \multicolumn{1}{c}{40000} &
  \multicolumn{1}{c}{\textit{Average $\Delta U$}} \\ \midrule
\multirow{2}{*}{\textbf{AAVI}} & \textit{DQN} & 24.7 & -17.5 & -63.4 & -10.6 & -16.7\\
 & \textit{CQL} & -2.8 & 10.5 & -9.5 & 213.3 & \textbf{52.9}\\ \\
\multirow{2}{*}{CAVI} & \textit{DQN} & 8.9 & -9.3 & -24.6 & -48.0 & -12.2\\
 & \textit{CQL} & -8.8 & 0.5 & 4.4 & 1.1 & -0.7\\ \\
\multirow{2}{*}{AI} & \textit{DQN} & -377.3 & -399.3 & -386.8 & -388.5 & -388.0\\
 & \textit{CQL} & -25.6 & 235.3 & -88.2 & -49.9 & 17.9\\ \\
\multirow{2}{*}{SSI} & \textit{DQN} & 265.7 & -408.2 & -334.9 & -396.3 & -218.4\\
 & \textit{CQL} & 35.8 & 4.0 & 11.9 & -22.8 & 7.2\\ \bottomrule
\end{tabular}
\end{table*}

%% file: sections/5-conclusion.tex
\section{Conclusions}
Data efficiency in the offline learning of MDPs is highly coveted. Exploiting the intuition of an expert about the nature of the model can help to learn dynamics that better represent reality.

In this work, we built a semi-automated tool that can aid an expert in providing a statistical data-driven validation of her/his intuition about some properties of the environment. Correct deployment of the tool could improve the performance of the optimal policy obtained by solving the learned MDP.
Indeed, our results suggest that the proposed algorithm can effectively detect a symmetry of the dynamics of an MDP with high accuracy and that exploiting this knowledge can not only reduce the distributional shift but also provide performance gain in an envisaged optimal control of the system. However, when applied to ORL environments with DNN, all the prescriptions (and issues) about hyperparameter fine-tuning well known to ORL practitioners persist.

Besides its pros, the current work is still constrained by several limitations.
We note that the quality of the approach in continuous MDPs is greatly affected by the architecture of the Normalizing Flow used for Density Estimation and, more generally, by the state-action space preprocessing.
In detail, sometimes an environment is endowed by symmetries that an expert can not straightforwardly perceive in the default representation of the state-action space and a transformation would be required (imagine the very same CartPole, but with also the linear speed and position of the car expressed in polar coordinates).

In the future we plan: (i) to expand this approach by trying out more recent Normalizing Flow architectures like FFJORD \cite{DBLP:conf/iclr/GrathwohlCBSD19}; (ii) to consider combinations of multiple symmetries; (iii) after the offline detection of a symmetry, to exploit the data augmentation to improve the learning phase of online agents.